\documentclass[letterpaper, 10 pt, conference]{ieeeconf}  

\IEEEoverridecommandlockouts                              

\usepackage{epsfig} 
\usepackage{amsmath} 
\usepackage{amssymb}  
\usepackage{color}
\usepackage{cite}
\usepackage{hyperref}
\usepackage{booktabs}
\usepackage[caption=false,font=footnotesize]{subfig}
\usepackage[dvipsnames]{xcolor}
\usepackage{soul}
\newcounter{RNum}
\renewcommand{\theRNum}{\arabic{RNum}}
\newcommand{\Remark}{\noindent\textit{\textbf{Remark}~\refstepcounter{RNum}\textbf{\theRNum}: }}
\newcommand{\NoOne}[1]{\textcolor{red}{#1}}
\newcommand{\NoTwo}[1]{\textcolor{green}{#1}}
\newcommand{\NoThree}[1]{\textcolor{blue}{#1}}
\soulregister\NoOne7
\soulregister\NoTwo7
\soulregister\NoThree7
\soulregister\Remark7

\title{\LARGE \bf
Tracker Meets Night: A Transformer Enhancer for UAV Tracking
}

\author{Junjie Ye$^{1}$, Changhong Fu$^{1}$, Ziang Cao$^{2}$, Shan An$^{3}$, Guangze Zheng$^{1}$, and Bowen Li$^{1}$
\thanks{*Corresponding Author}
\thanks{$^{1}$Junjie Ye, Changhong Fu, Guangze Zheng, and Bowen Li are with the School of Mechanical Engineering, Tongji University, Shanghai 201804, China.
        {\tt\small changhongfu@tongji.edu.cn}}%
\thanks{$^{2}$Ziang Cao is with the School of Automotive Studies, Tongji University, Shanghai 201804, China.
		}%
\thanks{$^{3}$Shan An is with Tech \& Data Center, JD.COM Inc., Beijing 100108, China.}
}

\begin{document}

\maketitle
\thispagestyle{empty}
\pagestyle{empty}

\begin{abstract}

Most previous progress in object tracking is realized in daytime scenes with favorable illumination. State-of-the-arts can hardly carry on their superiority at night so far, thereby considerably blocking the broadening of visual tracking-related unmanned aerial vehicle (UAV) applications. 
To realize reliable UAV tracking at night, a spatial-channel Transformer-based low-light enhancer (namely SCT), which is trained in a novel task-inspired manner, is proposed and plugged prior to tracking approaches.
To achieve semantic-level low-light enhancement targeting the high-level task, the novel spatial-channel attention module is proposed to model global information while preserving local context.
In the enhancement process, SCT denoises and illuminates nighttime images simultaneously through a robust non-linear curve projection.
Moreover, to provide a comprehensive evaluation, we construct a challenging nighttime tracking benchmark, namely DarkTrack2021, which contains 110 challenging sequences with over 100 K frames in total. 
Evaluations on both the public UAVDark135 benchmark and the newly constructed DarkTrack2021 benchmark show that the task-inspired design enables SCT with significant performance gains for nighttime UAV tracking compared with other top-ranked low-light enhancers.
Real-world tests on a typical UAV platform further verify the practicability of the proposed approach.
The DarkTrack2021 benchmark and the code of the proposed approach are publicly available at \url{https://github.com/vision4robotics/SCT}.

\end{abstract}

\section{INTRODUCTION}

Visual tracking is a fundamental task in numerous unmanned aerial vehicle (UAV)-based applications, \textit{e.g.}, target following~\cite{Li2021RAL}, autonomous landing~\cite{Javier2021RAL}, and self-localization~\cite{Ye2021TIE}. Given an initial position of an object, trackers are expected to estimate the location of the object in the following period. Recent years have witnessed a huge leap forward in visual tracking. Continuously emerging tracking approaches~\cite{Li2019CVPR, Bhat2019ICCV,  Danelljan2020CVPR, Cao2021ICCV, Chen2021CVPR} keep setting state-of-the-arts (SOTAs) in large-scale benchmarks. Nevertheless, previous progress is made on daytime sequences captured under favorable illumination conditions. In real-world applications of UAVs, vision systems are always required to provide robust performance around the clock, while recent studies~\cite{li2021allday, Ye2021IROS} show that SOTA trackers can hardly maintain their superiority in low-light conditions. \textit{Hence, the broadening of related UAV applications is impeded heavily by unstable tracking performance at nighttime.}

Generally, images captured at night are always with dim brightness and low-intensity contrast. In this case, feature extractors trained with daytime images lose their effectiveness and lead to unsatisfying object tracking. Along with poor illuminance, the high-level noise also damages the structural details of images, further degrading tracking performance. \textit{What is worse, there are few public nighttime tracking benchmarks with full annotations that are large enough to afford tracker training.}

\begin{figure}[!t]	
	\centering
	\includegraphics[width=0.99\linewidth]{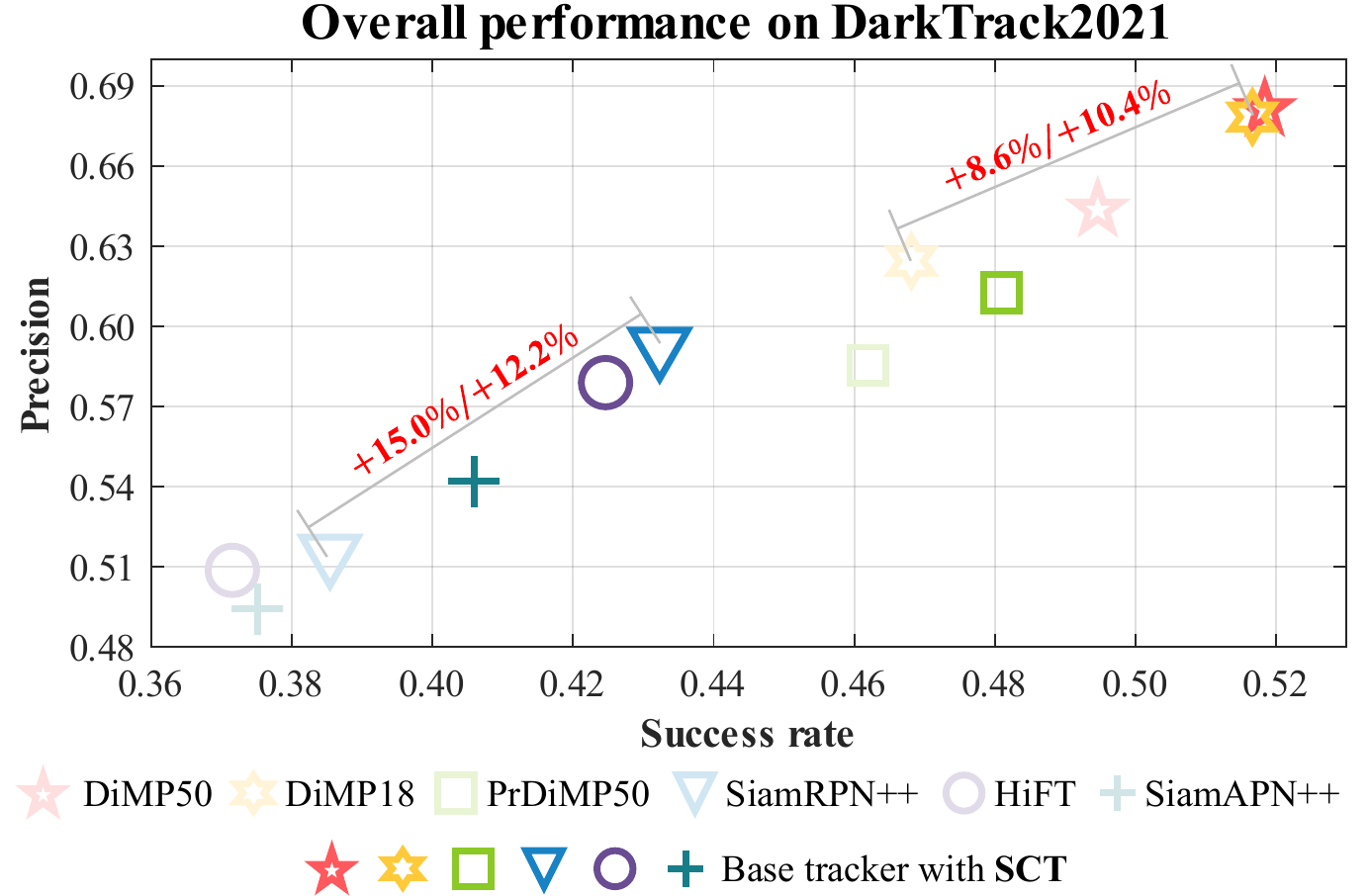}
	\caption
	{
		Overall performance of SOTA trackers~\cite{Li2019CVPR, Bhat2019ICCV, Danelljan2020CVPR, Cao2021IROS, Cao2021ICCV} with the proposed SCT enabled (markers in a \textcolor{darkgray}{\textbf{dark}} color) or not (markers in a \textcolor{lightgray}{\textbf{light}} color) in the newly constructed nighttime UAV tracking benchmark---DarkTrack2021. SCT significantly boosts the nighttime tracking performance of trackers in a plug-and-play manner.
	}
	\label{fig:fig1}
\end{figure}


Intuitively, one may turn to low-light enhancement approaches and serve them as a preprocessing step. However, most existing low-light enhancers~\cite{Guo2017TIP, Li2021TPAMI, Jiang2021TIP, Zhang2021CVPR, liu2021ruas} are with the purpose of facilitating human perception. The metrics in low-light enhancement, \textit{e.g.}, the peak signal-to-noise ratio (PSNR) and the structural similarity (SSIM) are designed for signal fidelity, which is not fully aligned with that of high-level tasks. Therefore, their effectiveness in high-level tasks is unsatisfying, and there is still a large margin for improvement~\cite{Liang2021TMM, Ye2021IROS}. Another disadvantage that impedes them from directly being deployed on UAVs is the computation complexity. Due to the limited computational resource, algorithms with a high computational burden are unaffordable on UAVs.

In visual tracking, the framework always consists of a backbone network and a task head~\cite{Bertinetto2016ECCVW, Li2019CVPR}. 
The performance of object tracking is heavily dependent on the effectiveness of feature extraction.
Naturally, to carry on trackers' favorable performance at daytime to low-light conditions, a promising solution is to narrow the gap between the features from low-light images and normal light ones.
Therefore, we adopt a task-inspired perceptual loss. 
Adopting the tracking backbone as a guideline, the objective of training a low-light enhancer is to minimize the difference between features extracted from enhanced images and those from normal light images. 
In this way, the low-light enhancer is trained to meet the requirement of object tracking instead of human perception.
Moreover, the proposed approach does not need any annotated night tracking sequences for training. Only a small set of paired low/normal light images is sufficient to bring trackers to nighttime.

\begin{figure*}[!t]	
	\centering
	\includegraphics[width=0.99\linewidth]{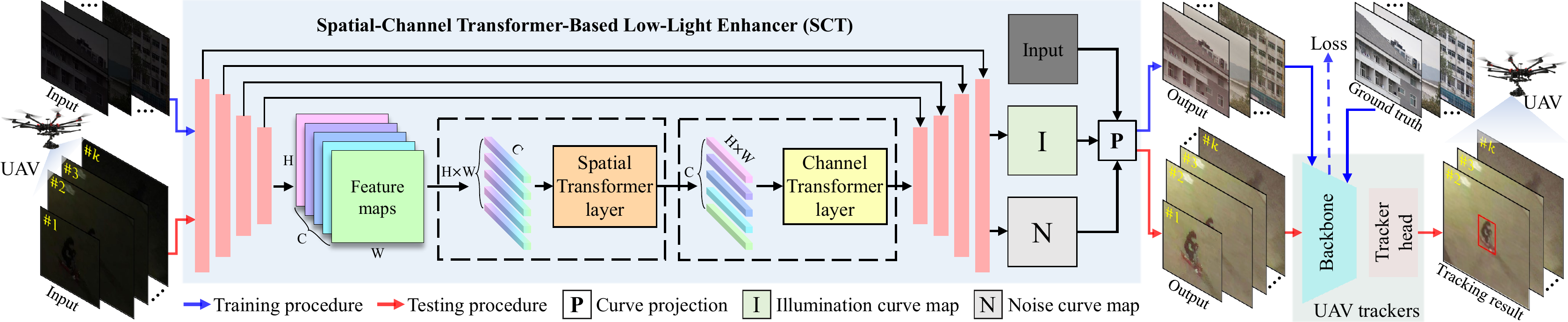}
	\caption
	{
		Overview of our nighttime UAV tracking pipeline. The spatial-channel Transformer-based low-light enhancer (SCT) firstly estimates the illumination and noise curve maps from the input. Through the robust curve projection, low-light input is then illuminated. With a task-inspired perceptual loss, SCT is trained to meet the requirement of the feature extractor in visual tracking. In the testing procedure, SCT is plugged prior to a tracking network to enhance the template and searching patches, therefore bringing favorable gains for trackers.
	}
	\label{fig:main}
\end{figure*}

Recent studies~\cite{Jiang2020CVPR, Zamir2021CVPR} have shown the effectiveness of attention modules in image restoration tasks. Since low-light enhancement in this work devotes to facilitating the high-level perception task rather than solely pixel-level illuminance adjustment, the modeling of global information is relatively more crucial. Due to the inherent locality of convolution operations, convolutional neural networks (CNNs), however, struggle to model long-range inter-dependencies~\cite{Wang2018CVPR}.
Drawing lessons from the robust performance of the global self-attention in long-range relation modeling~\cite{vaswani2017nips, dosovitskiy2020image, Liu2021swin}, we construct our enhancer with a spatial-channel Transformer-based attention module, dubbed SCT. 
To conserve local context in the meantime, the feed-forward network (FFN) of the standard Transformer is substituted by a residual convolutional block. 
In contrast to image-to-image mapping, we consider image enhancement as a non-linear curve projection task following~\cite{Li2021TPAMI}. As we target real-world nighttime UAV applications, the inevitable noise frequently damages the structural details of captured images. We consequently design a curve projection model with a noise term.
An overall performance comparison shown in Fig.~\ref{fig:fig1} demonstrates that SCT considerably assists nighttime tracking.
Figure~\ref{fig:main} illustrates the workflow of the whole framework. 
Prior to the visual tracking network, image enhancement is performed. The proposed SCT firstly estimates illumination and noise parameter maps, and the robust curve projection formulation is then adopted to adjust the low-light template patch and search patches.

Moreover, to facilitate the development of nighttime UAV tracking and provide a comprehensive evaluation, this work further collects and constructs a nighttime UAV tracking benchmark, namely DarkTrack2021. Consisting of 110 well-annotated high definition sequences with over 100 K frames in total, DarkTrack2021 involves numerous objects and scenes. 

The contributions of this paper lie in four-fold:
\begin{itemize}
	\item We propose a task-inspired low-light enhancer namely SCT. Trained in a task-tailored manner, SCT serves well in real-world nighttime UAV tracking.
	\item We construct a low-light enhancer with a spatial-channel Transformer and a robust non-linear curve projection model to achieve favorable low-light image enhancement.
	\item We collect a nighttime UAV tracking benchmark, namely DarkTrack2021, to provide comprehensive and detailed evaluations for nighttime UAV tracking.
	\item Evaluations on both the public UAVDark135~\cite{li2021allday} and newly constructed DarkTrack2021 benchmarks demonstrate the effectiveness of SCT in facilitating nighttime UAV tracking. Real-world tests on a typical UAV platform further verify its efficiency and practicability.
\end{itemize}

\section{Related work}
\subsection{Low-Light Image Enhancement}
Towards meeting the requirement of human visual perception, low-light image enhancement has attracted wide attention for a long time. C.~Guo \textit{et al.}~\cite{Guo2017TIP} propose to estimate and refine the illumination map for restoring the input following the Retinex model~\cite{Edwin1977retinex}. To overcome the lack of sufficient paired low/normal light data, Y. Jiang \textit{et al.}~\cite{Jiang2021TIP} train a generative adversarial network (GAN) in an unsupervised manner. More recently, R. Liu \textit{et al.}~\cite{liu2021ruas} introduce the neural architecture search into low-light enhancement to construct a lightweight and effective structure. Reformulating the low-light enhancement task as a non-linear curve projection problem, DCE++~\cite{Li2021TPAMI} estimates a parameter map from a low-light input and adopts a curve projection model to illuminate low-light images iteratively. However, such a curve projection model focus on retouching illumination while neglecting the noise that inevitably appears in real-world nighttime imaging. Despite the visually pleasing enhancement results, the effectiveness of SOTA enhancers in facilitating nighttime tracking is still far from satisfying~\cite{Ye2021IROS}. This can be partly attributed to the unalignment of their optimization goals with visual tracking.
\subsection{Vision Transformers}
In the past few years, Transformer~\cite{vaswani2017nips} has shown extraordinary performance in the field of natural language processing (NLP). Due to the ability to effectively capture long-range inter-independencies, recent studies introduce Transfomer into computer vision tasks~\cite{dosovitskiy2020image, Liu2021swin, Cao2021ICCV, Zheng_2021_CVPR}. However, Transformer for low-level tasks, \textit{e.g.}, image restoration, is not well investigated so far. H. Chen \textit{et al.}~\cite{Chen_2021_CVPR} pre-train a model comprised of several standard Transformer layers with multiple convolutional heads and tails to accomplish different tasks, while it suffers from high computational burn and large-scale training data. Z. Wang \textit{et al.}~\cite{wang2021uformer} construct a Transformer-based U-shaped model to realize several image restoration tasks, \textit{i.e.}, image denoising, deraining, deblurring, and demoiréing. In contrast, this work introduces a Transformer-based spatial-channel attention module to model global inter-dependencies and exploit global context for effective low-light enhancement.

\subsection{Visual Perception at Nighttime}

Visual perception tasks at night have attracted a lot of attention. Several works\cite{Sakaridis2019ICCV, Wu2021CVPR, Sakaridis2020TPAMI} pertain to semantic segmentation focus on transferring daytime models to nighttime. In\cite{Sasagawa2020ECCV}, a low-light enhancement model and an object detection model are merged to realize nighttime object detection. However, object tracking at nighttime is not well investigated relative to other computer vision tasks. Unsatisfying tracking performance at nighttime still critically hinders the broadening of UAV applications up to the present. Motivated by the insight, B. Li \textit{et al.}\cite{li2021allday} incorporate a low-light enhancer into a correlation filter-based tracker while is lack of flexibility and failing to utilize robust deep features. J.~Ye \textit{et al.}\cite{Ye2021IROS} attempt to boost nighttime tracking with a Retinex-inspired enhancer, \textit{i.e.}, DarkLighter. Nevertheless, DarkLighter is designed intuitively and suffers from weak collaboration with visual tracking. To this concern, this work proposes to learn an enhancer in a task-inspired manner, thus yielding promising effectiveness in nighttime tracking.

\section{Proposed Method}
This work proposes a novel task-inspired low-light enhancer to facilitate nighttime UAV tracking. As shown in Fig.~\ref{fig:main}, the proposed SCT involves a Transformer-based spatial-channel attention module to estimate the illumination and noise curve maps of the input. A low-light image is then enlightened through a robust non-linear projection. 
The training of SCT is guided by the feature extractor of object tracking to realize tracking-tailored low-light enhancement.
In the testing stage, SCT illuminates the template and searching patches for trackers.
\subsection{Spatial-Channel Transformer-Based Low-Light Enhancer}
In contrast to UNet~\cite{Ronneberger2015MICCAI} with pure convolutional layers, SCT adopts a U-shaped CNN-Transformer hybrid structure. The introduction of Transformer-based spatial-channel attention enables SCT a better perception ability of global information. Given a low-light image $\mathbf{X}_{\rm i} \in \mathbb{R}^{3 \times H_{\rm i} \times W_{\rm i}}$ with the resolution of $H_{\rm i} \times W_{\rm i}$ pixels, $K$ CNN encoders are utilized to generate the feature maps $\mathbf{X} \in \mathbb{R}^{C \times H \times W}$. Each encoder consists of two $3\times3$ convolutional layers with a LeakyReLU activation function, followed by a downsampling layer. After each encoder stage, the channels of feature maps are doubled, and the spatial resolution is halved, except for the first encoder that generates a feature map of 32 dimensions. Therefore, $H=\frac{H_{\rm i}}{2^K}$, and $W=\frac{W_{\rm i}}{2^K}$. To achieve spatial-wise attention, we flatten $\mathbf{X}$ in spatial dimension to $\mathbf{X_{\rm s}}\in \mathbb{R}^{(H \times W) \times C }$ and input it into the spatial Transformer layer. $\mathbf{X_{\rm s}}$ can be regarded as $H \times W$ feature vectors with a length of $C$. Each vector corresponds to a specific spatial position of $C$ feature channels. Next, the obtained feature maps are flattened in the channel dimension, obtaining $\mathbf{X_{\rm c}}\in \mathbb{R}^{C \times (H \times W)}$. In this way, each feature vector involves the spatial information among all $H \times W$ pixels of the corresponding channel. Through the spatial-channel attention module, the correlations among spatial positions and channels are captured. 
Following the attention module are $K$ CNN decoders mirrored to the encoders. The last convolutional layer is followed by a Tanh activation function, which outputs an illumination curve map $\mathbf{I}\in \mathbb{R}^{3\times H_{\rm i} \times W_{\rm i}}$ and a noise curve map $\mathbf{N}\in \mathbb{R}^{3\times H_{\rm i} \times W_{\rm i}}$.
Utilizing the proposed robust curve projection, the enlightened image $\mathbf{X}_{\rm o}\in \mathbb{R}^{3\times H_{\rm i} \times W_{\rm i}}$ is obtained. Below we introduce the Transformer layer and the robust curve projection in detail.

\subsubsection{Transformer layer} Since SCT serves as a preprocessing step of nighttime tracking in resource-limited UAVs, it is supposed to be efficient and effective. In consideration of the high computational burden of the vanilla Transformer~\cite{vaswani2017nips}, we adopt the non-overlapping window-based multi-head self-attention (W-MSA) following~\cite{dosovitskiy2020image, Liu2021swin}. For instance, given a flattened feature $\mathbf{F} \in \mathbb{R}^{N \times L}$, $\mathbf{F}$ is first reshaped to 2D shape as $\mathbf{F} \in \mathbb{R}^{N \times \sqrt{L} \times \sqrt{L}}$, and then evenly partitioned to non-overlapping windows with a size of $M \times M$. Next, each patch is flattened and transposed to $\mathbf{f}_j \in \mathbb{R}^{M^2 \times N}$. The subscript $j$ denotes the $j$-th window. Therefore, $\mathbf{F}$ consists of $\frac{L}{M^2}$ patches:
\begin{equation}\label{equ:partition}
	\mathbf{F} = \left\{\mathbf{f}_1, \mathbf{f}_2, ..., \mathbf{f}_j,...,\mathbf{f}_{\frac{L}{M^2}}\right\}
	\quad.
\end{equation}

Following Fig.~\ref{fig:resffn} (a), self-attention is then performed on each patch. The computation of the Transformer layer can be formulated as:
\begin{equation}\label{equ:transformer}
	\begin{split}
		\hat{\mathbf{F}}'=&{\rm W\text{-}MSA} ({\rm LN}(\mathbf{F})) + \mathbf{F}\quad,\\
		\mathbf{F}'=&{\rm MLP} ({\rm LN}(\hat{\mathbf{F}}')) + \hat{\mathbf{F}}'\quad,\\
	\end{split}
\end{equation} 
where $\hat{\mathbf{F}}'$ and $\mathbf{F}'$ denote the output of W-MSA and the multi-layer perception (MLP), respectively. LN is layer normalization. Inspired by~\cite{Liu2021swin}, relative position bias $\mathbf{B}\in\mathbb{R}^{M^2 \times M^2}$ is adopted into the self-attention. Therefore, the calculation of attention in each window can be represented as:
\begin{equation}\label{equ:attention}
	{\rm Attention}(\mathbf{Q}, \mathbf{K}, \mathbf{V}) = {\rm SoftMax}(\frac{\mathbf{Q}\mathbf{K}^\mathsf{T}}{\sqrt{d}}+\mathbf{B})\mathbf{V}
	\quad,
\end{equation}
where $\mathbf{Q}, \mathbf{K}, \mathbf{V}\in \mathbf{R}^{M^2 \times d}$ represent the query, key, and value matrixs. $d$ is the dimension of the query.

\begin{figure}[!t]	
	\centering
	\includegraphics[width=0.75\linewidth]{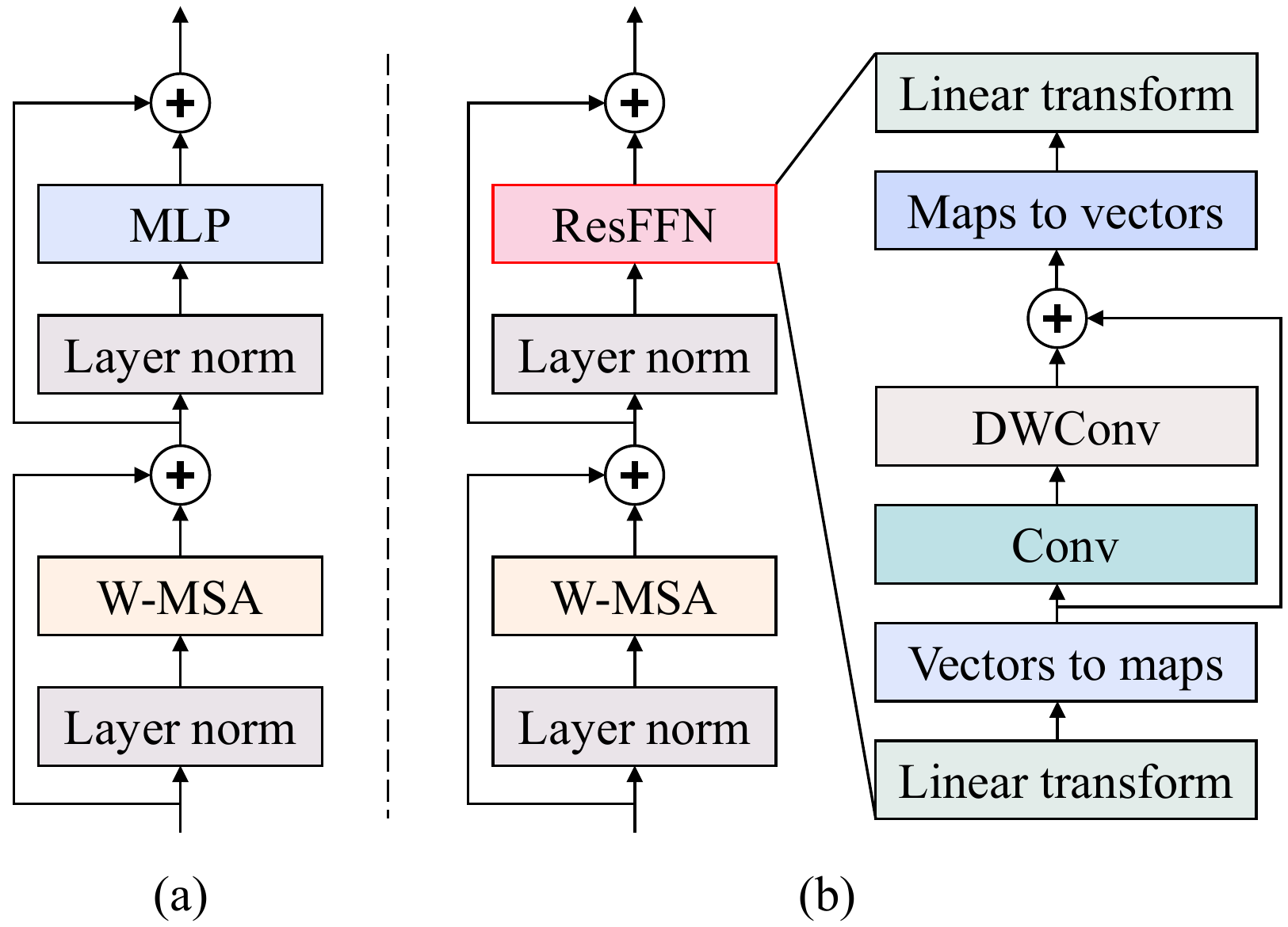}
	\caption
	{
		Illustration of the original window-based Transformer layer (a) and our redesigned Transformer layer with ResFFN (b). The abbreviations MLP and W-MSA denote the multi-layer perception and the window-based multi-head self-attention. Attributing to the introduction of ResFFN, local detail information is enhanced.
	}
	\label{fig:resffn}
\end{figure}

Moreover, previous studies~\cite{li2021localvit, wu2021cvt} show that standard Transformer presents drawbacks in leveraging local details, while local context is crucial to image enhancement tasks. Hence, we replace MLP in the original Transformer layer with a residual convolution feed-forward network (ResFFN), aiming to make it up by convolution layers' favorable local context perception ability. Figure~\ref{fig:resffn} (b) illustrates the structure of ResFFN, where the input gets through a linear transform layer and is reshaped to a 2D shape, followed by a residual operation involving a convolutional layer (Conv) and a depthwise convolutional layer (DWConv). The output is then generated by a flattening operation and a linear transform. In this way, the local context in the features is preserved and enhanced. Consequently, the attention operation is reformulated as:
\begin{equation}\label{equ:transformer_res}
	\begin{split}
		\hat{\mathbf{F}}'&={\rm W\text{-}MSA} ({\rm LN}(\mathbf{F})) + \mathbf{F}\quad,\\
		\mathbf{F}'&={\rm ResFFN} ({\rm LN}(\hat{\mathbf{F}}')) + \hat{\mathbf{F}}'\quad.\\
	\end{split}
\end{equation}

\subsubsection{Robust curve projection} In~\cite{Li2021TPAMI}, the low-light enhancement task is reformulated as a non-linear curve projection problem. Estimating a light curve parameter map $\mathbf{I}$ from a low-light image $\mathbf{X}_{\rm i}$, the final output of the enhanced image $\mathbf{X}_{\rm o}$ can be obtained following $T$ iterative projections:
\begin{equation}\label{equ:projection}
	\begin{split}
		\mathbf{X}_{\rm i}^t=\mathbf{X}_{\rm i}^{t-1} + \mathbf{I} \odot \mathbf{X}_{\rm i}^{t-1} \odot &(1-\mathbf{X}_{\rm i}^{t-1}), \quad t=1,...,T ,\\
		\mathbf{X}_{\rm o}=&\mathbf{X}_{\rm i}^T \quad,\\
	\end{split}
\end{equation} 
where $\mathbf{X}_{\rm i}^t$ is the intermediate results of enhancing process, and $\mathbf{X}_{\rm i}^0=\mathbf{X}_{\rm i}$. $\odot$ denotes element-wise multiplication.

However, in this curve projection model, noise is not taken into consideration. To realize illuminance retouching for normal pixels and denoising for noise pixels simultaneously, we redesign the curve projection model and involve a noise term:
\begin{equation}\label{equ:projection_noise}
	\begin{split}
		\hat{\mathbf{X}}_{\rm i}^{t-1} = \mathbf{X}&_{\rm i}^{t-1} -\mathbf{N} \quad,\\
		\mathbf{X}_{\rm i}^t=\hat{\mathbf{X}}_{\rm i}^{t-1} + \mathbf{I} \odot \hat{\mathbf{X}}_{\rm i}^{t-1} \odot &(1-\hat{\mathbf{X}}_{\rm i}^{t-1}), \quad t=1,...,T ,\\
		\mathbf{X}_{\rm o}=&\mathbf{X}_{\rm i}^T \quad.\\
	\end{split}
\end{equation} 

In every iteration, the input $\mathbf{X}_{\rm i}^{t-1}$ first subtracts the noise map estimated by SCT to obtain $\hat{\mathbf{X}}_{\rm i}^{t-1}$ for denoising, and then undergoes the curve projection for illuminance retouching.

\Remark The utilization of the noise term achieves dedicated denoise and makes the following curve projection focus more on illuminance retouching, thus yielding a pleasant enhancement.

\subsection{Task-Inspired Training}
Since the purpose of low-light enhancement in this work is to promote UAV tracking, the proposed SCT enhancer is supposed to meet the requirement of trackers. In light of that feature extraction is essential for high-level tasks, this work proposes to let the backbone $\mathcal{F}$ in trackers guide the training process of low-light enhancement. Hence, a task-inspired loss $\mathcal{L}$ is formulated as:
\begin{equation}\label{equ:loss}
	\mathcal{L} =  \sum_m\frac{1}{c_m h_m w_m}\left\Vert \mathcal{F}_m(\mathbf{X}_{\rm o}) - \mathcal{F}_m(\mathbf{Y}) \right\Vert_2^2
	\quad ,
\end{equation} 
where $c_m$, $h_m$, $w_m$ represent corresponding dimensions of feature maps from the $m$-th layer, and $\mathbf{Y}$ denotes the ground truth normal light image. 
In our implementation, the widely adopted modified AlexNet~\cite{Bertinetto2016ECCVW} is employed as the loss backbone $\mathcal{F}$, for the reason that knowledge in a general backbone can reflect the requirement of object tracking for features and its relatively shallow structure ensures an easy convergence.
The 3rd, 4th, and 5th layers are used for loss calculation since trackers generally utilize features from these layers. Thus, $m=3,4,5$. 

\Remark Trained with this task-inspired loss only, the experiments show that the learned SCT enhancer realizes favorable gains in nighttime tracking compared to other SOTA low-light enhancers.

\subsection{SCT for Nighttime UAV Tracking}

\begin{figure}[!t]

	\centering	
	\includegraphics[width=0.98\linewidth]{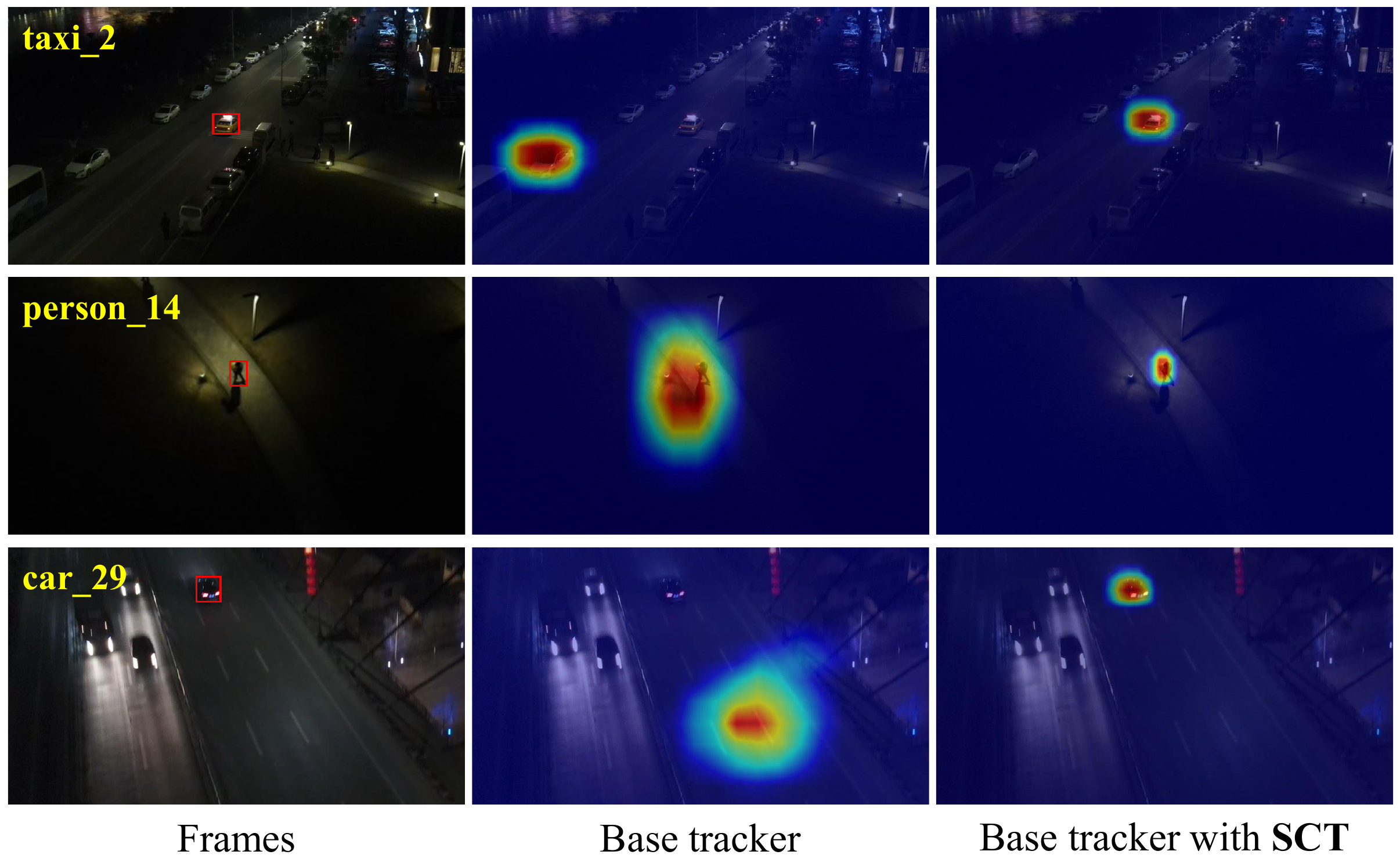}
	\caption
	{
		Visual comparison of confidence maps generated from the base tracker with SCT enabled or not. The \NoOne{red} boxes in the first column mark tracked objects. The images are from the proposed DarkTrack2021 benchmark, with sequence names displayed on the top left corner of the original frames. The base tracker loses the efficiency of locating objects in the darkness, while SCT raises its perception ability favorably.
	}
	\label{fig:simi}
\end{figure}

As shown in Fig.~\ref{fig:main}, SCT is plugged prior to tracking approaches. Template (or searching) patches are firstly cropped from newly captured low-light frames according to the initial position (or the estimated position of the last frame). Before being fed into the feature extractor, the template and searching patches are illuminated by SCT. Next, the tracker head exploits features extracted by the backbone to calculate confidence maps of the template patch and the current searching patch, and update the position of the object. A visual comparison of confidence maps generated from a base tracker with SCT activated or not is illustrated in Fig.~\ref{fig:simi}. It can be seen clearly that with the enhancement of SCT to the template/searching patches, the base tracker yields satisfying perception ability of objects in darkness.

\Remark Through our manner, visual trackers are extended to the night without any need for nighttime tracking sequences. Only a small set of paired low/normal light images is sufficient for the training of SCT. Simply plug it in front of the backbone network, SCT can shed light on the darkness and promote nighttime UAV tracking. 

\section{The DarkTrack2021 Benchmark}
A nighttime UAV tracking benchmark, namely DarkTrack2021, is constructed in this work for comprehensive evaluations. Compared with the existing benchmark~\cite{li2021allday} for nighttime UAV tracking in literature, scenes in the newly developed benchmark are generally captured in complex urban from a higher altitude, where light conditions are more cluttered, frequently bringing severe illumination variation and overexposure/underexposure challenges. We hope the developed DarkTrack2021 can promote the development of nighttime aerial tracking for the community.

\begin{figure}[!t]
	\centering
	\includegraphics[width=0.98\linewidth]{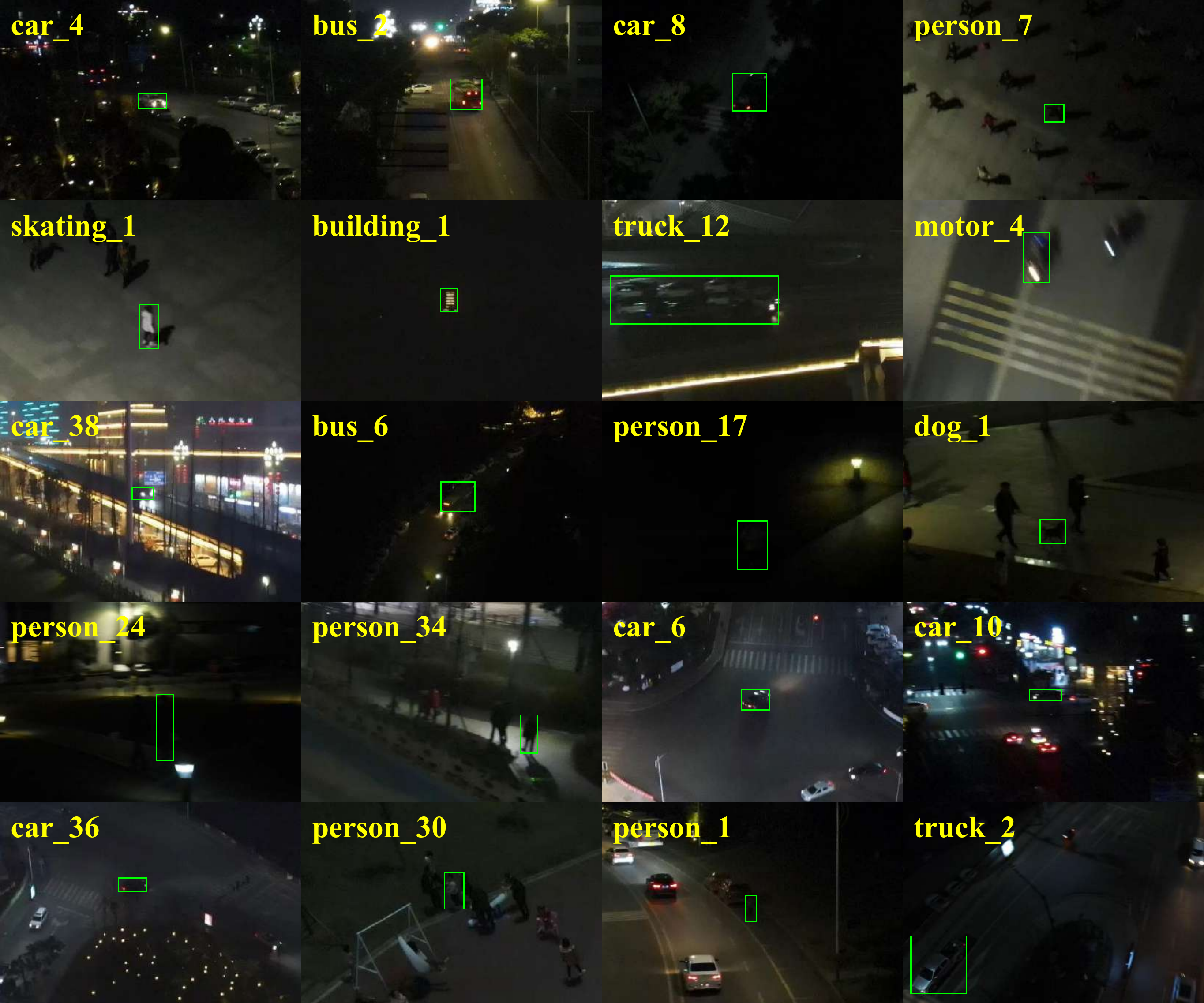}
			\setlength{\abovecaptionskip}{0pt}
	\caption{First frames of some selected sequences from DarkTrack2021. The \NoTwo{green} boxes mark the tracking objects, while the top left corner of the images displays sequence names. Low brightness makes it hard to identify objects, which leads nighttime UAV tracking to an extremely challenging task. Best viewed on-screen with high-resolution.}
	\label{fig:benchmark}
\end{figure}

\noindent\textbf{Sequence collection} All sequences are captured at nighttime in urban scenes by the authors, with utmost effort to design various challenging tracking scenarios. We adopt the DJI Mavic Air 2 UAV\footnote{More information of the UAV can be found at \url{https://www.dji.com/cn/mavic-air-2}.}, with a frame rate of 30 frames/s (FPS). Finally, the benchmark comprises 110 challenging sequences with 100,377 frames in total. The shortest, longest, and the average length of sequences are respectively 92, 6579, and 913 frames. Figure~\ref{fig:benchmark} displays some first frames of selected sequences. The objects in DarkTrack2021 contain person, bus, car, truck, motor, dog, building, \textit{etc.}, covering abundant scenarios of real-world UAV nighttime tracking tasks. Large amounts of scenarios with various challenges, including viewpoint change, fast motion, large occlusion, low resolution, low brightness, out-of-view, \textit{etc.}, are involved in DarkTrack2021. In that case, DarkTrack2021 provides extensive nighttime tracking performance evaluations.

\noindent\textbf{Annotation} Apart from sequence collection, delicate annotating is conducted for comprehensive evaluation. All frames in DarkTrack2021 are manually annotated by annotators familiar with object tracking, under the principle of tightly surrounding objects' edge pixels with bounding boxes. After the first annotating process, visual inspection by the authors and annotation refinement by annotators are performed iteratively until all frames are annotated with high quality.

\section{Experiments}
In this section, a comprehensive ablation study is performed to verify components in SCT. A comparison of performance in facilitating nighttime UAV tracking is then conducted among SCT and other SOTA low-light enhancers. To testify the generalizability of SCT, it is further implemented on other SOTA trackers. Finally, real-world tests are conducted to verify the applicability of the proposed approach. The newly constructed DarkTrack2021 benchmark, the source code of SCT, along with some demo videos are made available at \url{https://github.com/vision4robotics/SCT}.

\subsection{Implementation Details}
The total number $K$ of CNN encoders/decoders is set to 4. Inspired by~\cite{Li2021TPAMI}, the input images are first scaled to 128$\times$128 to perform curve maps estimation for speeding up. There is respectively one Transformer layer for spatial attention and channel attention. The window size is 4$\times$4. We employ 485 paired low/normal light images from the LOL dataset~\cite{Chen2018Retinex} for training. The AdamW~\cite{loshchilov2019decoupled} is utilized as our optimizer. The initial learning rate is set to 0.0008, with a weight decay of 0.02. The training is warmed up with 5 epochs and lasts a total of 100 epochs. The batch size is set to 32. The training patch size is adopted as 256$\times$256, with a random crop operation on the original training image. 
Experiments are conducted on a PC with an Intel i9-9920X CPU, an NVIDIA TITAN RTX GPU, and a 32GB RAM. Finally, an NVIDIA Jetson AGX Xavier serves as the real-world test platform to testify the practicability of SCT in nighttime UAV tracking. 

\subsection{Evaluation Metrics} 
Since we target nighttime UAV tracking, evaluation metrics in visual tracking are adopted to rank the performance instead of those in low-light enhancement. The experiments follow the one-pass evaluation (OPE)~\cite{Mueller2016ECCV}, which involves two metrics, respectively precision and success rate. The center location error (CLE) of the predicted position and the ground truth position of the object is utilized to calculate the precision. The percentage of the frames with a CLE below a given threshold is presented as the precision plot (PP). As generally adopted, the threshold of 20 pixels is used to rank trackers. In addition, the success rate is measured by the intersection over union (IoU) of the estimated bounding box and the ground truth one. The percentage of the frames whose IoU is greater than a preset maximum threshold makes up the success plot (SP). In general, the area-under-the-curve (AUC) on SP is used to rank the success rate of the trackers.

\begin{table}[!b]
	\scriptsize
	\centering
	\caption{Comparison of SCT with different modules enabled on UAVDark135. The best results are highlighted in \NoOne{red} font. $\Delta$~represents the percentages exceeding the baseline.}
	\begin{tabular}{cccccc}
		\toprule
		\multicolumn{1}{c}{SA} & \multicolumn{1}{c}{CA} & \multicolumn{1}{c}{ResFF} & \multicolumn{1}{c}{Denoise} & Succ./Prec. & $\Delta$ (\%) \\
		\midrule
		\multicolumn{1}{c}{\checkmark} & \multicolumn{1}{c}{\checkmark} & \multicolumn{1}{c}{\checkmark} & \multicolumn{1}{c}{\checkmark} & \textcolor[rgb]{ 1,  0,  0}{\textbf{0.421/0.547}} & \textcolor[rgb]{ 1,  0,  0}{\textbf{+13.3/+15.4}} \\
		\midrule
		\multicolumn{1}{c}{\checkmark} &        & \multicolumn{1}{c}{\checkmark} & \multicolumn{1}{c}{\checkmark} & 0.415/0.536 & +11.7/+13.0 \\
		& \multicolumn{1}{c}{\checkmark} & \multicolumn{1}{c}{\checkmark} & \multicolumn{1}{c}{\checkmark} & 0.409/0.523 & +10.1/+10.3 \\
		&        &        & \multicolumn{1}{c}{\checkmark} & 0.412/0.533 & +10.9/+12.4 \\
		\midrule
		\multicolumn{1}{c}{\checkmark} & \multicolumn{1}{c}{\checkmark} &        & \multicolumn{1}{c}{\checkmark} & 0.401/0.514 & +8.0/+8.5 \\
		\midrule
		\multicolumn{1}{c}{\checkmark} & \multicolumn{1}{c}{\checkmark} & \multicolumn{1}{c}{\checkmark} &        & 0.393/0.505 & +5.6/+6.6 \\
		\midrule
		&        &        &        & 0.372/0.474 & -/- \\
		\bottomrule
	\end{tabular}
	\label{tab:abla}%
\end{table}%

\subsection{Ablation Study}
This subsection investigates the performance of different variants of SCT. A typical Siamese tracker---SiamRPN++~\cite{Li2019CVPR} with the AlexNet backbone, is utilized as the baseline. All variants are trained in the same way. Tracking results on UAVDark135~\cite{li2021allday} are reported in TABLE~\ref{tab:abla}. SA, CA, and denoise denote the spatial attention, the channel attention, and the noise term, respectively. The bottom row indicates the original baseline without enhancement, which shows an unsatisfying performance, while the introduction of complete SCT brings gains of \textbf{13.3}\% and \textbf{15.4}\% in success rate and precision, respectively. 
\subsubsection{Attention module} Ablating CA and SA respectively, the gains of SCT bringing to tracking performance degraded to some extend, validating the effectiveness of both the spatial attention and the channel attention. As shown in the fourth line of TABLE~\ref{tab:abla}, replacing the spatial-channel Transformer with a general CNN bottleneck, the pure CNN UNet enhancer brings an increase of 10.9\% in success rate and 12.4\% in precision, which is $\sim$2.5\% lower than that the complete SCT brought. One can conclude that the proposed spatial-channel Transformer-based structure performs better than the original CNN UNet.
\subsubsection{ResFF} As shown in the fifth line of TABLE~\ref{tab:abla}, adopting MLP rather than our proposed ResFF as FFN, the baseline only gains a performance of 8.0\% in success rate and 8.5\% in precision. SCT with ResFF surpasses that with MLP in a large margin. The effectiveness of ResFF in preserving and enhancing local context can be verified, which is crucial for both low-light enhancement and tracking.
\subsubsection{Noise term} Removing the noise term of the robust curve projection model, the benefit of introducing SCT is halved, which shows the importance of appropriate denoising in real-world nighttime UAV tracking.

\Remark Deactivating the key modules in SCT, the learned enhancer can still promote nighttime tracking in certain, which is attributed to the task-inspired training.

\subsection{Comparison with SOTA Low-Light Enhancers}
\begin{figure}[!t]	
	\centering
	\includegraphics[width=0.99\linewidth]{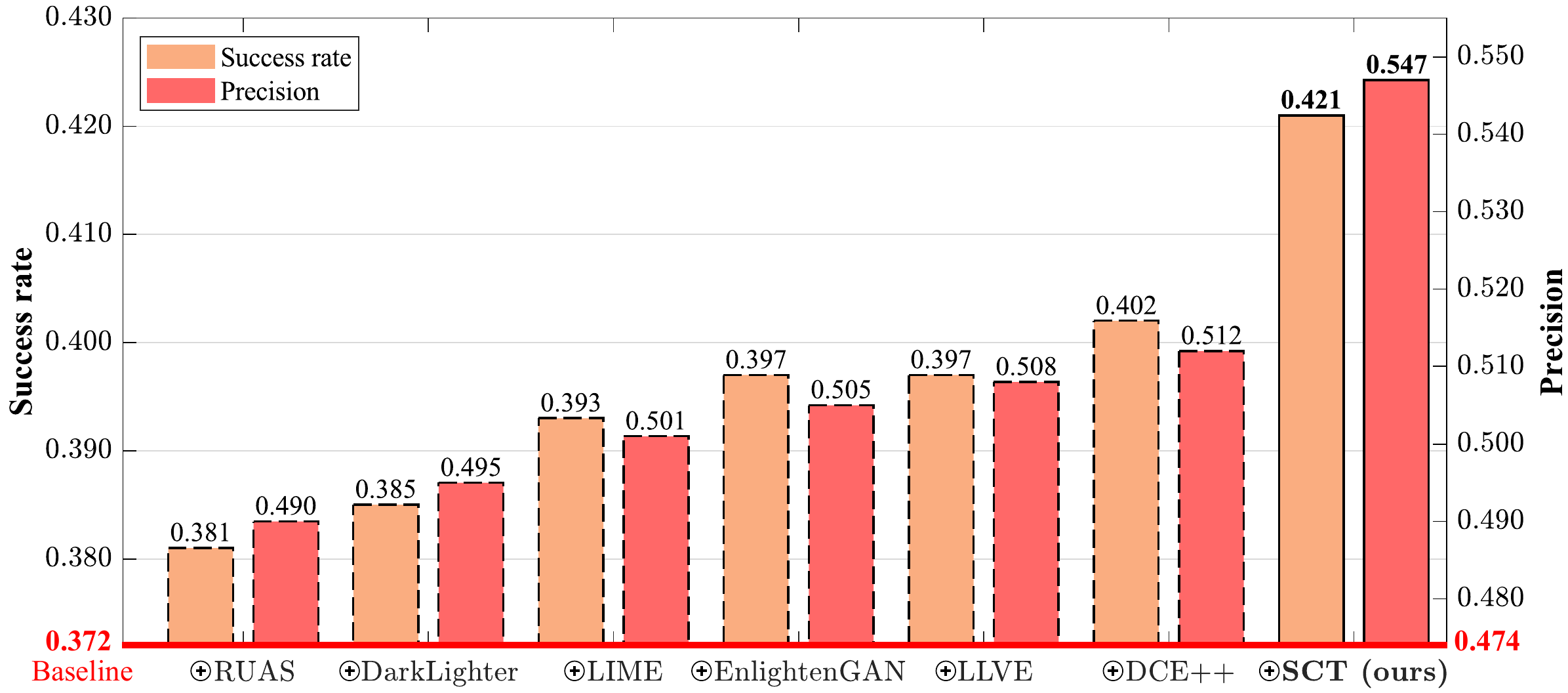}
	\caption
	{
		Tracking performance comparison with different low-light enhancers implemented. The bottom \NoOne{red} line denotes baseline without any enhancement. The proposed SCT raises the baseline by a large margin.
	}
	\label{fig:enhancers}
\end{figure}

To present the advantage of SCT in facilitating nighttime UAV tracking, it is further compared with other 6 SOTA low-light enhancers, including DCE++~\cite{Li2021TPAMI}, EnlightenGAN~\cite{Jiang2021TIP}, LIME~\cite{Guo2017TIP}, LLVE~\cite{Zhang2021CVPR}, RUAS~\cite{liu2021ruas}, and DarkLighter~\cite{Ye2021IROS}. The tracking performances of the baseline with different low-light enhancement approaches enabled on the UAVDark135 benchmark are displayed in Fig.~\ref{fig:enhancers}. The results demonstrate that the task-specific design makes SCT perform surprisingly in facilitating nighttime tracking, while other general low-light enhancers are at a similar suboptimal level. The proposed approach raises the baseline by over \textbf{13}\% and \textbf{15}\% in success rate and precision, respectively, far surpassing the second-best DCE++, which gains a promotion of around 8\%. Though DarkLighter is designed with the motivation of UAV tracking, the benefits it brings are inferior due to the weak collaboration with the tracking task.

\begin{figure*}[!t]	
	\centering
	\subfloat[Results on UAVDark135.]
	{
		\includegraphics[width=0.243\linewidth]{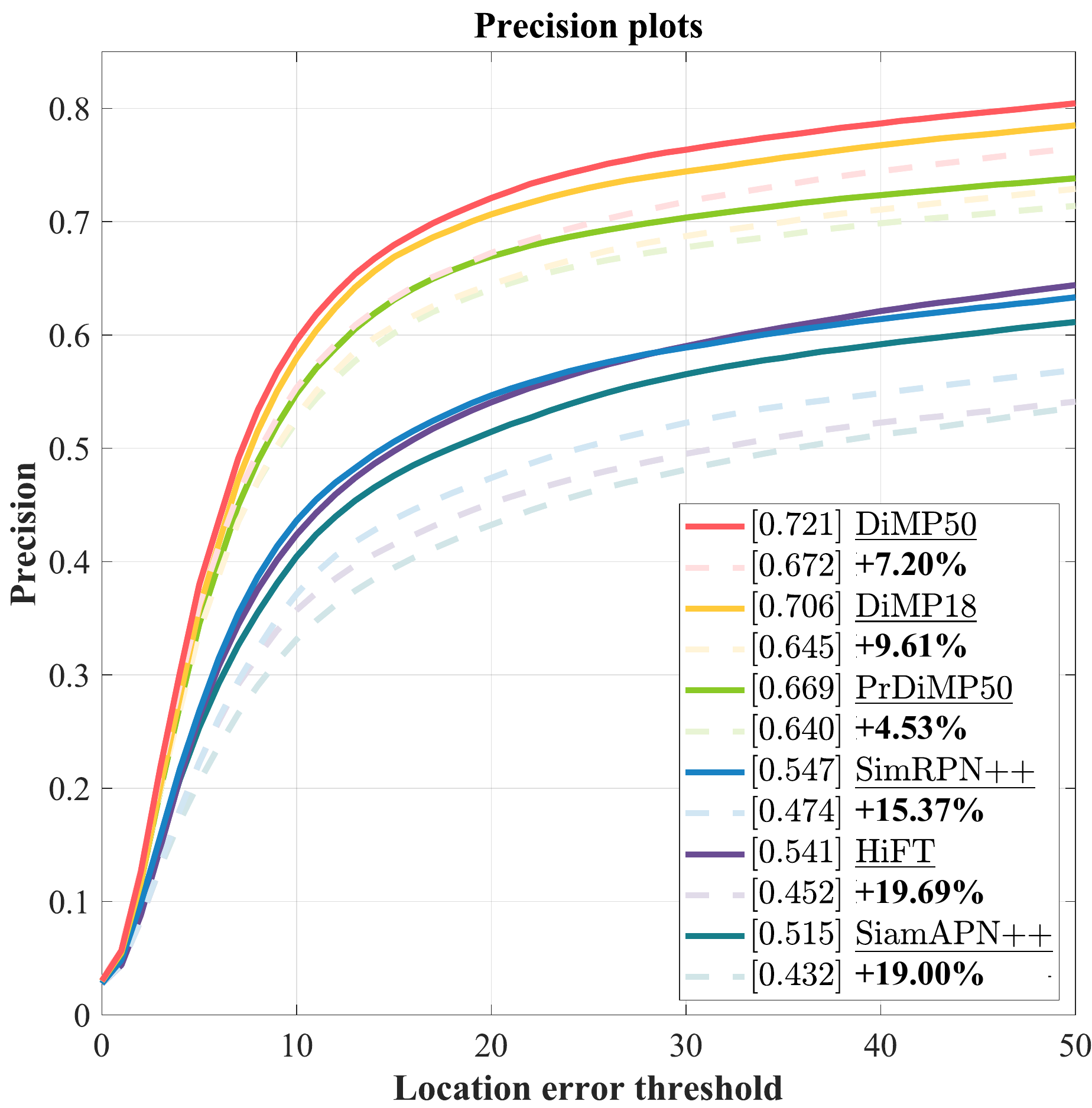}
		\includegraphics[width=0.243\linewidth]{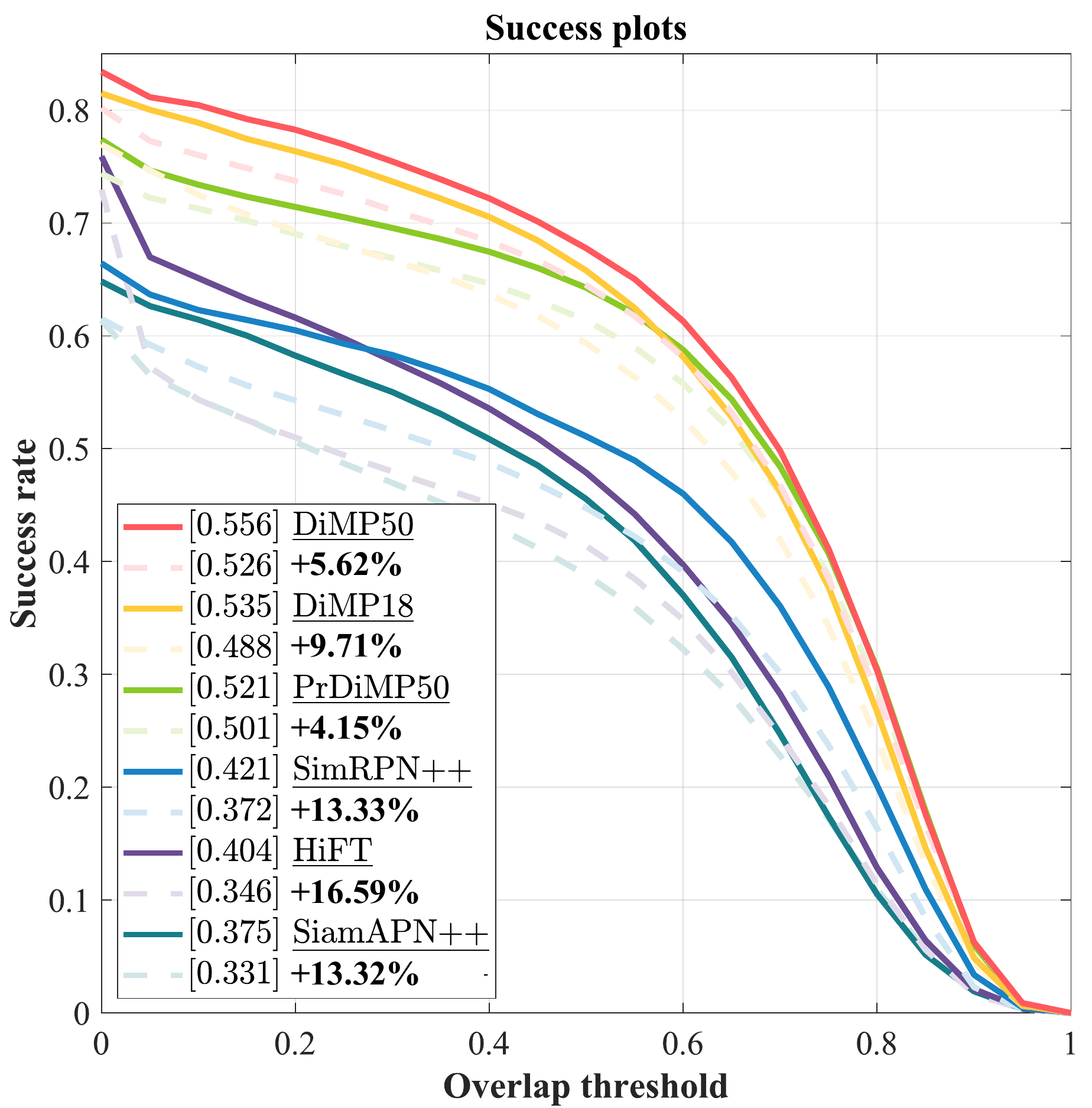}
	}
	\subfloat[Results on DarkTrack2021.]
	{\includegraphics[width=0.243\linewidth]{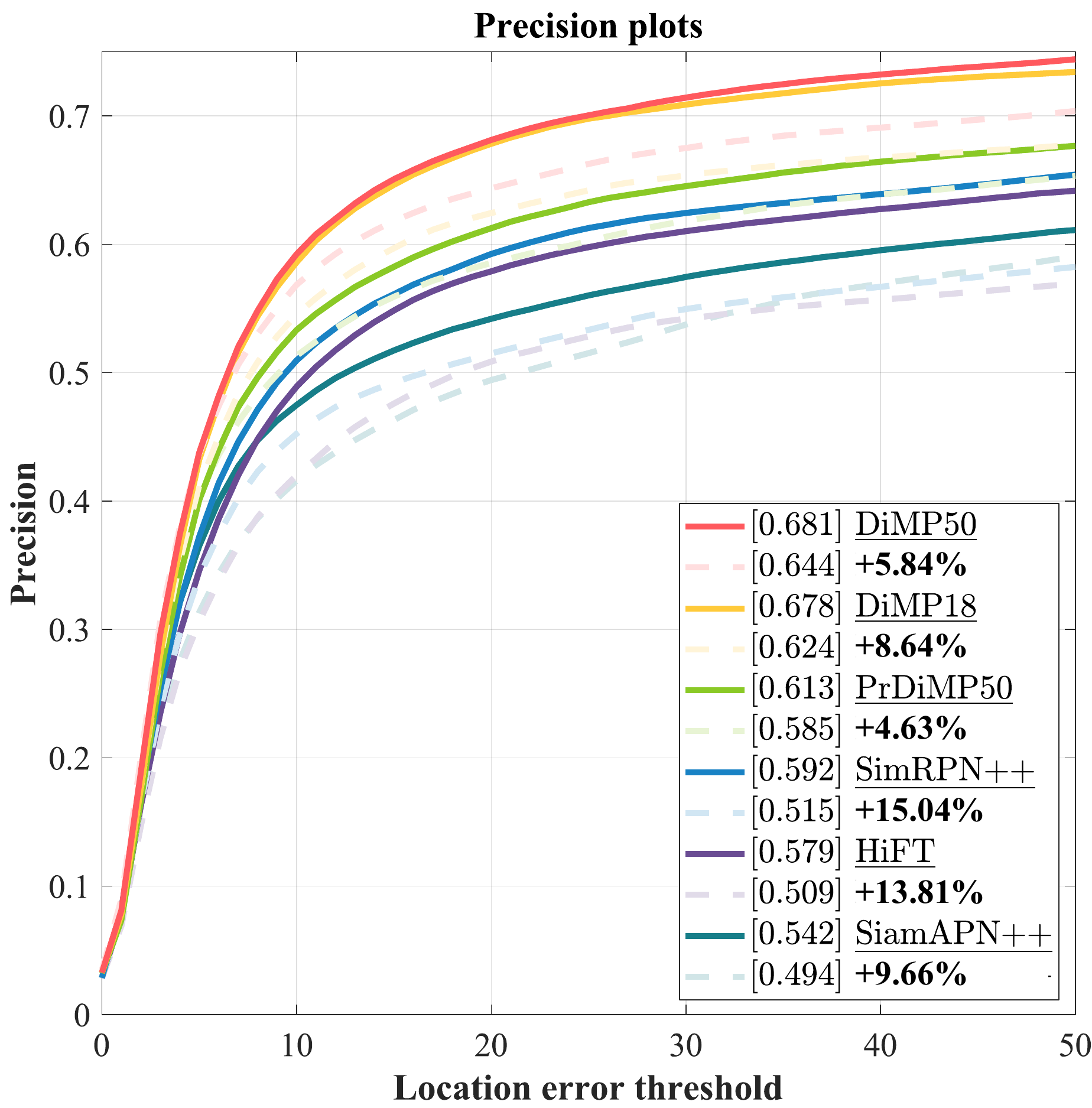}
		\includegraphics[width=0.243\linewidth]{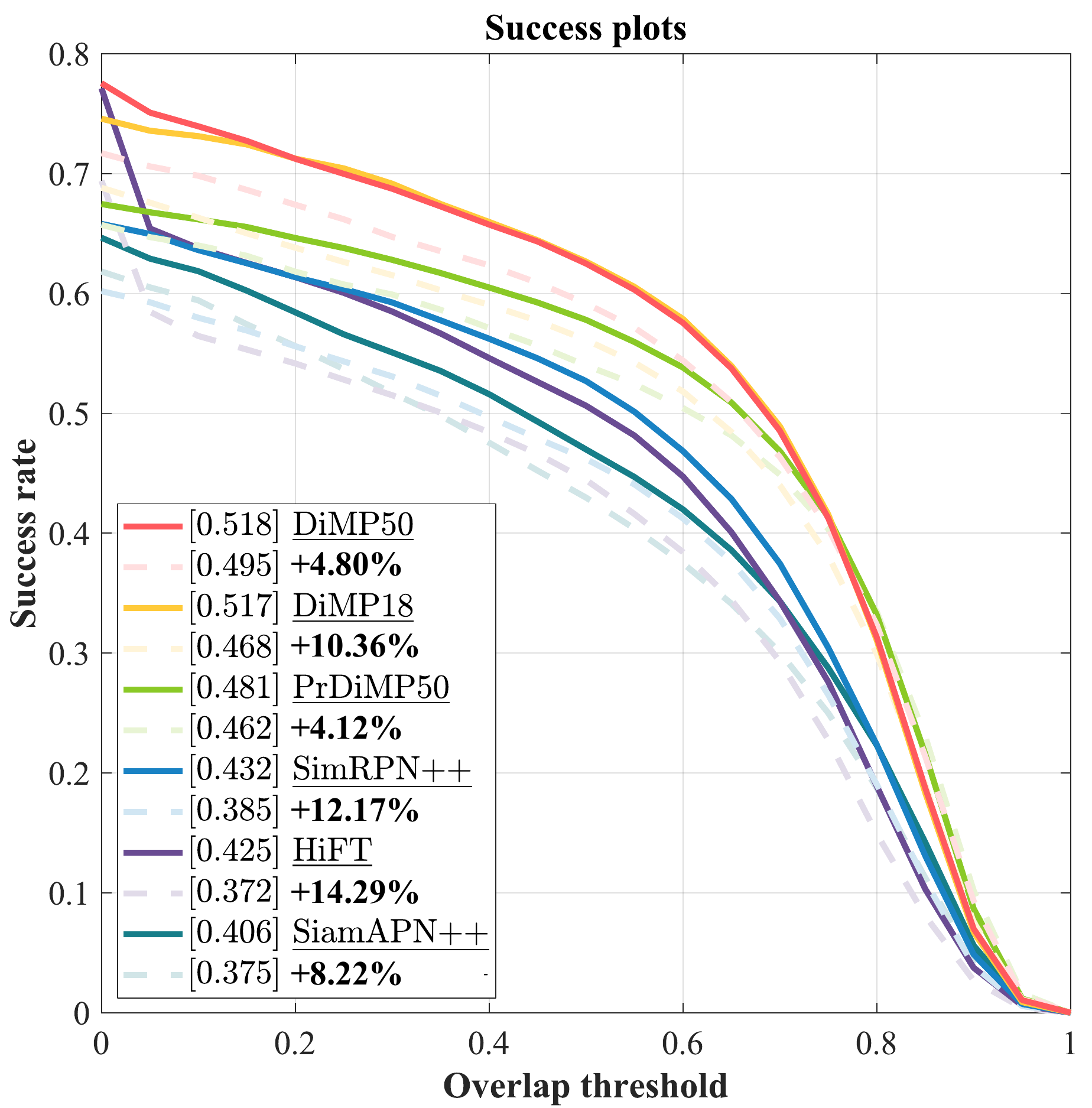}}
	\caption
	{
		Overall performance of SOTA trackers with SCT activated (plots with deep colors) or not (plots with light colors). Trackers underlined in the legend denote tracking performance with the activation of the proposed SCT, while the percentages report the performance gains brought by SCT. The results show the proposed enhancement module considerably promotes the performance of all involved trackers.
	}
	\label{fig:all}
\end{figure*}

\subsection{SCT in SOTA Trackers}
To testify the performance gain of SCT in different trackers, the proposed approach is further implemented in 6 SOTA trackers, including AlexNet-based trackers (SiamRPN++~\cite{Li2019CVPR}, SiamAPN++~\cite{Cao2021IROS}, and HiFT~\cite{Cao2021ICCV}), ResNet18-based tracker (DiMP18~\cite{Bhat2019ICCV}), and ResNet50-based trackers (DiMP50~\cite{Bhat2019ICCV} and PrDiMP50~\cite{Danelljan2020CVPR}). Figure~\ref{fig:all} shows the overall performance of trackers on UAVDark135 and our newly constructed DarkTrack2021 with SCT enabled or not. The performance gains of each tracker are reported in the legend. The results show that SCT significantly boosts the nighttime tracking performance of all involved trackers on both benchmarks. For instance, in UAVDark135, with the facilitating of SCT, HiFT boosts \textbf{19.69}\% and \textbf{16.59}\% in precision and success rate, respectively.  In DarkTrack2021, SCT helps DiMP50 earn promotions of \textbf{5.84}\% and \textbf{4.80}\%, reaching 0.681 and 0.518 in precision and success rate, respectively. In addition, we found that the gains of SCT brought partly depend on the depth of backbones. The shallow ones earn more ($>$13\% in most AlexNet-based trackers), while the deep ones earn less ($\sim$5\% in ResNet50-based trackers). We conjecture that shallow backbones lose feature extraction efficiency more severely at night, thus benefiting more from low-light enhancement.
\begin{figure}[!t]
	\centering	
	\includegraphics[width=0.99\linewidth]{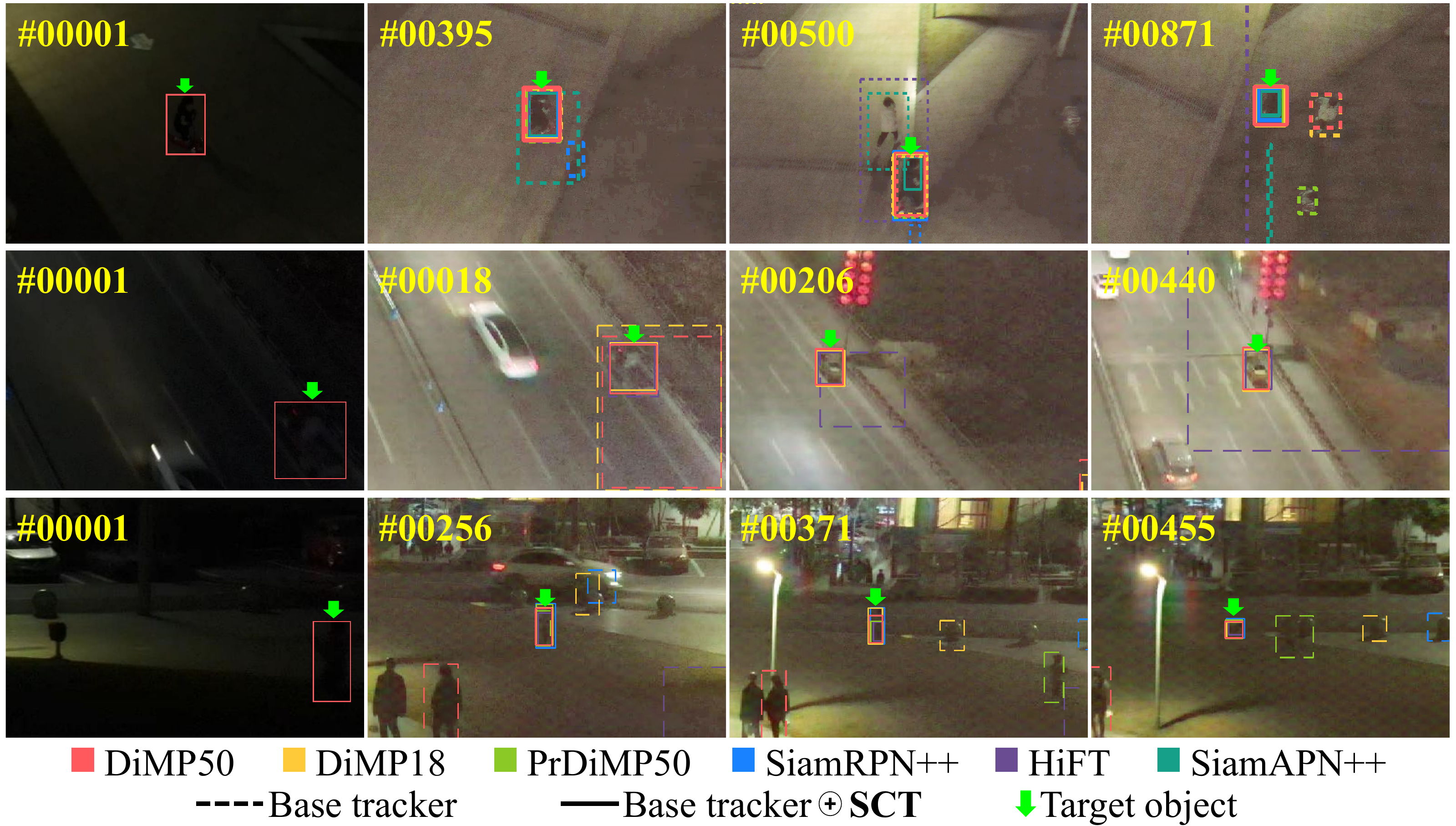}
	\caption
	{
		Qualitative results of trackers with SCT activated (solid boxes) or not (dashed boxes). The images are globally enhanced by SCT for visualization except for the first column. From top to down, the sequences are $person18$, $car23$, and $person24$ from the newly constructed DarkTrack2021. SCT boosts the nighttime tracking performance of trackers significantly.
	}
	\label{fig:qua}
\end{figure}

In Fig.~\ref{fig:qua}, some tracking screenshots are presented. Images are globally enhanced by SCT except for the first column for visualization, while only the template and search patches are enhanced in the practical tracking pipeline. We can conclude that SCT benefits the perception ability of trackers at night, thus raising tracking reliability considerably. 

\Remark Apart from favorable nighttime tracking performance gains, SCT also shows competitive image enhancement results.

\subsection{Real-World Tests}
\begin{figure}[!t]

	\centering	
	\includegraphics[width=0.99\linewidth]{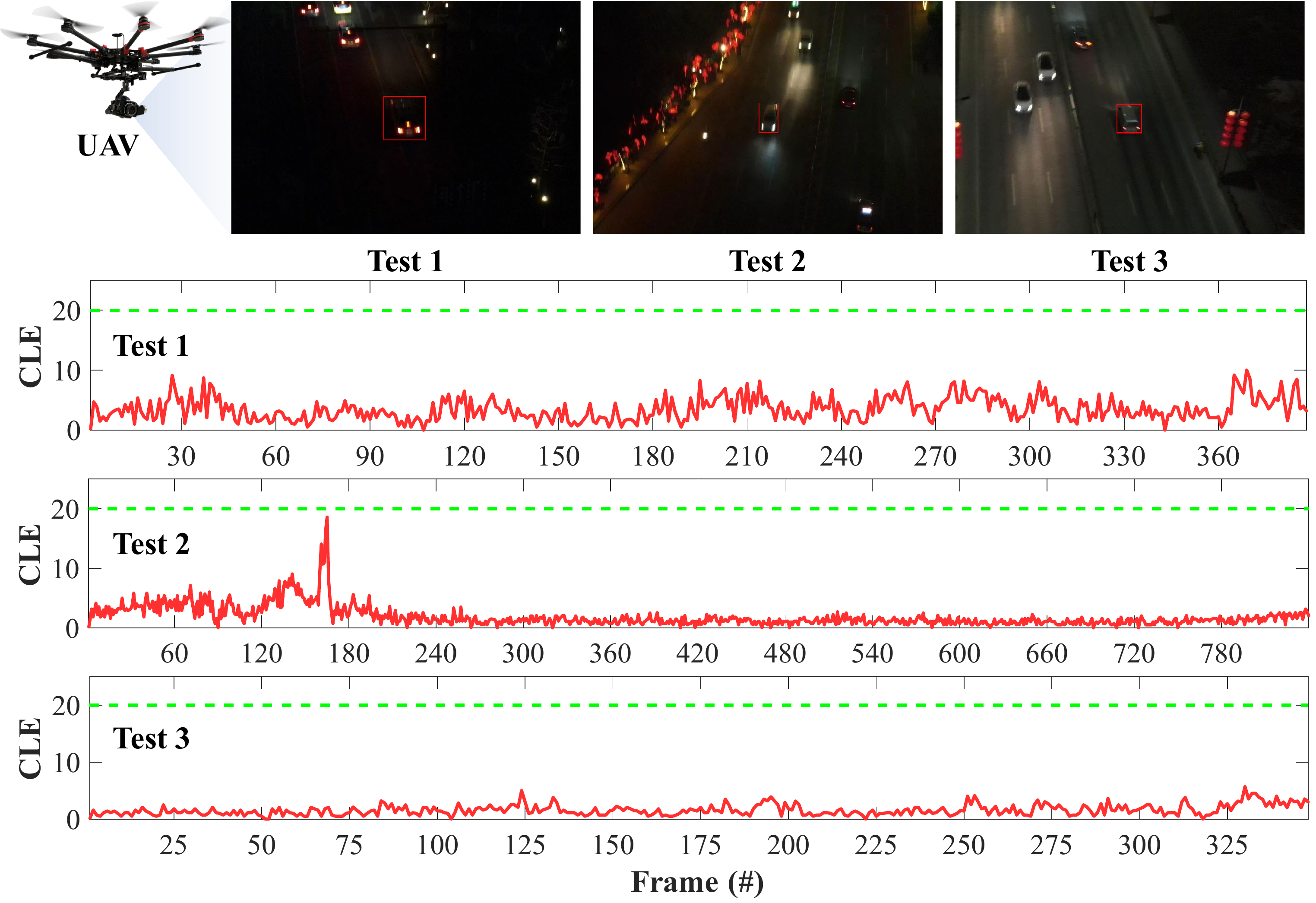}
	\caption
	{
		Real-world tests on a typical UAV platform. \NoOne{Red} bounding boxes denote the estimated positions. CLE curves between predictions and ground truth are drawn below. The \NoTwo{green} dashed line locates a threshold of 20 pixels, tracking errors within which are normally regarded as satisfying. The base tracker realizes favorable nighttime tracking assisted by SCT.
	}
	\label{fig:real}
\end{figure}

To demonstrate the applicability of SCT in real-world nighttime UAV tracking applications, we further adopt it on a typical embedded system, \textit{i.e.}, NVIDIA Jetson AGX Xavier. The results demonstrate that SCT realizes a promising real-time speed of $\sim$31.25 FPS without TensorRT acceleration. Further, Fig.~\ref{fig:real} shows some real-world nighttime tracking tests and CLE curves. The main challenges in the tests are low brightness, illumination variation, and fast motion. The CLE curves show that the prediction errors are within 20 pixels, which can be regarded as reliable tracking. With the assistance of SCT, the base tracker is able to effectively extract discriminative features and realize robust object localization at night.

\section{Conclusion}
This work proposes a task-inspired low-light enhancement approach, namely SCT. Adopting a novel spatial-channel Transformer module and a robust curve projection model, the proposed approach puts the requirement of visual tracking into great consideration, thus yielding a tracking-tailored enhancement performance. Working in a plug-and-play manner, SCT can be utilized as a unified method to boost nighttime UAV tracking. Moreover, a well-annotated nighttime UAV tracking benchmark is constructed for comprehensive tracking performance evaluation. The promising performance of SCT in benchmarks and real-world tests verifies its effectiveness and practicability, with a promising real-time speed. To sum up, we strongly believe that both the proposed tracking-customized low-light enhancer and the constructed nighttime tracking benchmark will considerably contribute to nighttime UAV tracking and expand related UAV applications at night.  
\section*{ACKNOWLEDGMENT}
This work was supported in part by the Natural Science Foundation of Shanghai under Grant 20ZR1460100 and in part by the National Natural Science Foundation of China under Grant 62173249.


\bibliographystyle{IEEEtran}
\bibliography{icra2022}

\begin{thebibliography}{10}
\providecommand{\url}[1]{#1}
\csname url@samestyle\endcsname
\providecommand{\newblock}{\relax}
\providecommand{\bibinfo}[2]{#2}
\providecommand{\BIBentrySTDinterwordspacing}{\spaceskip=0pt\relax}
\providecommand{\BIBentryALTinterwordstretchfactor}{4}
\providecommand{\BIBentryALTinterwordspacing}{\spaceskip=\fontdimen2\font plus
\BIBentryALTinterwordstretchfactor\fontdimen3\font minus
  \fontdimen4\font\relax}
\providecommand{\BIBforeignlanguage}[2]{{%
\expandafter\ifx\csname l@#1\endcsname\relax
\typeout{** WARNING: IEEEtran.bst: No hyphenation pattern has been}%
\typeout{** loaded for the language `#1'. Using the pattern for}%
\typeout{** the default language instead.}%
\else
\language=\csname l@#1\endcsname
\fi
#2}}
\providecommand{\BIBdecl}{\relax}
\BIBdecl

\bibitem{Li2021RAL}
J.~Li, H.~Xie, K.~H. Low, J.~Yong, and B.~Li, ``{Image-Based Visual Servoing of
  Rotorcrafts to Planar Visual Targets of Arbitrary Orientation},'' \emph{IEEE
  Robot. Automat. Lett.}, vol.~6, no.~4, pp. 7861--7868, 2021.

\bibitem{Javier2021RAL}
J.~González-Trejo, D.~Mercado-Ravell, I.~Becerra, and R.~Murrieta-Cid, ``{On
  the Visual-Based Safe Landing of UAVs in Populated Areas: A Crucial Aspect
  for Urban Deployment},'' \emph{IEEE Robot. Automat. Lett.}, vol.~6, no.~4,
  pp. 7901--7908, 2021.

\bibitem{Ye2021TIE}
J.~Ye, C.~Fu, F.~Lin, F.~Ding, S.~An, and G.~Lu, ``Multi-regularized
  correlation filter for uav tracking and self-localization,'' \emph{IEEE
  Trans. Ind. Electron.}, vol.~69, no.~6, pp. 6004--6014, 2022.

\bibitem{Li2019CVPR}
B.~Li, W.~Wu, Q.~Wang, F.~Zhang, J.~Xing, and J.~Yan, ``{SiamRPN++: Evolution
  of Siamese Visual Tracking with Very Deep Networks},'' in \emph{Proc.
  IEEE/CVF Conf. Comput. Vis. Pattern Recognit.}, 2019, pp. 4277--4286.

\bibitem{Bhat2019ICCV}
G.~Bhat, M.~Danelljan, L.~Van~Gool, and R.~Timofte, ``{Learning Discriminative
  Model Prediction for Tracking},'' in \emph{Proc. IEEE/CVF Int. Conf. Comput.
  Vis.}, 2019, pp. 6181--6190.

\bibitem{Danelljan2020CVPR}
M.~Danelljan, L.~Van~Gool, and R.~Timofte, ``{Probabilistic Regression for
  Visual Tracking},'' in \emph{Proc. IEEE/CVF Conf. Comput. Vis. Pattern
  Recognit.}, 2020, pp. 7181--7190.

\bibitem{Cao2021ICCV}
Z.~Cao, C.~Fu, J.~Ye, B.~Li, and Y.~Li, ``{HiFT: Hierarchical Feature
  Transformer for Aerial Tracking},'' in \emph{Proc. IEEE/CVF Int. Conf.
  Comput. Vis.}, 2021, pp. 15\,457--15\,466.

\bibitem{Chen2021CVPR}
X.~Chen, B.~Yan, J.~Zhu, D.~Wang, X.~Yang, and H.~Lu, ``{Transformer
  Tracking},'' in \emph{Proc. IEEE/CVF Conf. Comput. Vis. Pattern Recognit.},
  2021, pp. 8122--8131.

\bibitem{li2021allday}
B.~Li, C.~Fu, F.~Ding, J.~Ye, and F.~Lin, ``{All-Day Object Tracking for
  Unmanned Aerial Vehicle},'' \emph{arXiv preprint arXiv:2101.08446}, pp.
  1--13, 2021.

\bibitem{Ye2021IROS}
J.~Ye, C.~Fu, G.~Zheng, Z.~Cao, and B.~Li, ``{DarkLighter: Light Up the
  Darkness for UAV Tracking},'' in \emph{Proc. IEEE/RSJ Int. Conf. Intell.
  Robots Syst.}, 2021, pp. 3079--3085.

\bibitem{Cao2021IROS}
Z.~Cao, C.~Fu, J.~Ye, B.~Li, and Y.~Li, ``{SiamAPN++: Siamese Attentional
  Aggregation Network for Real-Time UAV Tracking},'' in \emph{Proc. IEEE/RSJ
  Int. Conf. Intell. Robots Syst.}, 2021, pp. 3086--3092.

\bibitem{Guo2017TIP}
X.~Guo, Y.~Li, and H.~Ling, ``{LIME: Low-Light Image Enhancement via
  Illumination Map Estimation},'' \emph{IEEE Trans. Image Process.}, vol.~26,
  no.~2, pp. 982--993, 2017.

\bibitem{Li2021TPAMI}
C.~Li, C.~Guo, and C.~L. Chen, ``{Learning to Enhance Low-Light Image via
  Zero-Reference Deep Curve Estimation},'' \emph{IEEE Trans. Pattern Anal.
  Mach. Intell.}, pp. 1--14, 2021.

\bibitem{Jiang2021TIP}
Y.~Jiang, X.~Gong, D.~Liu, Y.~Cheng, C.~Fang, X.~Shen, J.~Yang, P.~Zhou, and
  Z.~Wang, ``{EnlightenGAN: Deep Light Enhancement Without Paired
  Supervision},'' \emph{IEEE Trans. Image Process.}, vol.~30, pp. 2340--2349,
  2021.

\bibitem{Zhang2021CVPR}
F.~Zhang, Y.~Li, S.~You, and Y.~Fu, ``{Learning Temporal Consistency for Low
  Light Video Enhancement from Single Images},'' in \emph{Proc. IEEE/CVF Conf.
  Comput. Vis. Pattern Recognit.}, 2021, pp. 4965--4974.

\bibitem{liu2021ruas}
R.~Liu, L.~Ma, J.~Zhang, X.~Fan, and Z.~Luo, ``{Retinex-inspired Unrolling with
  Cooperative Prior Architecture Search for Low-light Image Enhancement},'' in
  \emph{Proc. IEEE/CVF Conf. Comput. Vis. Pattern Recognit.}, 2021, pp.
  10\,556--10\,565.

\bibitem{Liang2021TMM}
J.~Liang, J.~Wang, Y.~Quan, T.~Chen, J.~Liu, H.~Ling, and Y.~Xu, ``{Recurrent
  Exposure Generation for Low-Light Face Detection},'' \emph{IEEE Trans.
  Multimedia}, pp. 1--14, 2021.

\bibitem{Bertinetto2016ECCVW}
L.~Bertinetto, J.~Valmadre, J.~F. Henriques, V.~Andrea, and P.~H.~S. Torr,
  ``{Fully-Convolutional Siamese Networks for Object Tracking},'' in
  \emph{Proc. Eur. Conf. Comput. Vis. Workshop}, 2016, pp. 850--865.

\bibitem{Jiang2020CVPR}
K.~Jiang, Z.~Wang, P.~Yi, C.~Chen, B.~Huang, Y.~Luo, J.~Ma, and J.~Jiang,
  ``{Multi-Scale Progressive Fusion Network for Single Image Deraining},'' in
  \emph{Proc. IEEE/CVF Conf. Comput. Vis. Pattern Recognit.}, 2020, pp.
  8343--8352.

\bibitem{Zamir2021CVPR}
S.~W. Zamir, A.~Arora, S.~Khan, M.~Hayat, F.~S. Khan, M.-H. Yang, and L.~Shao,
  ``{Multi-Stage Progressive Image Restoration},'' in \emph{Proc. IEEE/CVF
  Conf. Comput. Vis. Pattern Recognit.}, 2021, pp. 14\,816--14\,826.

\bibitem{Wang2018CVPR}
X.~Wang, R.~Girshick, A.~Gupta, and K.~He, ``{Non-local Neural Networks},'' in
  \emph{Proc. IEEE/CVF Conf. Comput. Vis. Pattern Recognit.}, 2018, pp.
  7794--7803.

\bibitem{vaswani2017nips}
A.~Vaswani, N.~Shazeer, N.~Parmar, J.~Uszkoreit, L.~Jones, A.~N. Gomez,
  {\L}.~Kaiser, and I.~Polosukhin, ``{Attention Is All You Need},'' in
  \emph{Proc. Adv. Neural Inf. Process. Syst.}, 2017, pp. 6000--6010.

\bibitem{dosovitskiy2020image}
A.~Dosovitskiy, L.~Beyer, A.~Kolesnikov, D.~Weissenborn, X.~Zhai \emph{et~al.},
  ``{An Image is Worth 16x16 Words: Transformers for Image Recognition at
  Scale},'' in \emph{Proc. Int. Conf. Learn. Representations}, 2021, pp. 1--12.

\bibitem{Liu2021swin}
Z.~Liu, Y.~Lin, Y.~Cao, H.~Hu, Y.~Wei, Z.~Zhang, S.~Lin, and B.~Guo, ``{Swin
  Transformer: Hierarchical Vision Transformer using Shifted Windows},'' in
  \emph{Proc. IEEE/CVF Int. Conf. Comput. Vis.}, 2021, pp. 10\,012--10\,022.

\bibitem{Edwin1977retinex}
E.~H. Land, ``{The Retinex Theory of Color Vision},'' \emph{Sci. Amer.}, vol.
  237, no.~6, pp. 108--129, 1977.

\bibitem{Zheng_2021_CVPR}
S.~Zheng, J.~Lu, H.~Zhao, X.~Zhu, Z.~Luo, Y.~Wang, Y.~Fu, J.~Feng, T.~Xiang,
  P.~H. Torr, and L.~Zhang, ``{Rethinking Semantic Segmentation from a
  Sequence-to-Sequence Perspective with Transformers},'' in \emph{Proc.
  IEEE/CVF Conf. Comput. Vis. Pattern Recognit.}, 2021, pp. 6877--6886.

\bibitem{Chen_2021_CVPR}
H.~Chen, Y.~Wang, T.~Guo, C.~Xu, Y.~Deng, Z.~Liu, S.~Ma, C.~Xu, C.~Xu, and
  W.~Gao, ``{Pre-Trained Image Processing Transformer},'' in \emph{Proc.
  IEEE/CVF Conf. Comput. Vis. Pattern Recognit.}, 2021, pp. 12\,294--12\,305.

\bibitem{wang2021uformer}
Z.~Wang, X.~Cun, J.~Bao, J.~Liu, and H.~Li, ``{Uformer: A General U-Shaped
  Transformer for Image Restoration},'' \emph{arXiv preprint arXiv:2106.03106},
  pp. 1--11, 2021.

\bibitem{Sakaridis2019ICCV}
C.~Sakaridis, D.~Dai, and L.~Van~Gool, ``{Guided Curriculum Model Adaptation
  and Uncertainty-Aware Evaluation for Semantic Nighttime Image
  Segmentation},'' in \emph{Proc. IEEE/CVF Int. Conf. Comput. Vis.}, 2019, pp.
  7373--7382.

\bibitem{Wu2021CVPR}
X.~Wu, Z.~Wu, H.~Guo, L.~Ju, and S.~Wang, ``{DANNet: A One-Stage Domain
  Adaptation Network for Unsupervised Nighttime Semantic Segmentation},'' in
  \emph{Proc. IEEE/CVF Conf. Comput. Vis. Pattern Recognit.}, 2021, pp.
  15\,764--15\,773.

\bibitem{Sakaridis2020TPAMI}
C.~Sakaridis, D.~Dai, and L.~Van~Gool, ``{Map-Guided Curriculum Domain
  Adaptation and Uncertainty-Aware Evaluation for Semantic Nighttime Image
  Segmentation},'' \emph{IEEE Trans. Pattern Anal. Mach. Intell.}, pp. 1--15,
  2020.

\bibitem{Sasagawa2020ECCV}
Y.~Sasagawa and H.~Nagahara, ``{YOLO in the Dark - Domain Adaptation Method for
  Merging Multiple Models},'' in \emph{Proc. Eur. Conf. Comput. Vis.}, 2020,
  pp. 345--359.

\bibitem{Ronneberger2015MICCAI}
O.~Ronneberger, P.~Fischer, and T.~Brox, ``{U-Net: Convolutional Networks for
  Biomedical Image Segmentation},'' in \emph{Proc. Int. Conf. Med. Image
  Comput. Comput.-Assist. Interv.}, 2015, pp. 234--241.

\bibitem{li2021localvit}
Y.~Li, K.~Zhang, J.~Cao, R.~Timofte, and L.~V. Gool, ``{LocalViT: Bringing
  Locality to Vision Transformers},'' \emph{arXiv preprint arXiv:2104.05707},
  pp. 1--10, 2021.

\bibitem{wu2021cvt}
H.~Wu, B.~Xiao, N.~Codella, M.~Liu, X.~Dai, L.~Yuan, and L.~Zhang, ``{CvT:
  Introducing Convolutions to Vision Transformers},'' in \emph{Proc. IEEE/CVF
  Int. Conf. Comput. Vis.}, 2021, pp. 22--31.

\bibitem{Chen2018Retinex}
C.~Wei, W.~Wang, W.~Yang, and J.~Liu, ``{Deep Retinex Decomposition for
  Low-Light Enhancement},'' in \emph{Proc. Brit. Mach. Vis. Conf.}, 2018, pp.
  1--12.

\bibitem{loshchilov2019decoupled}
I.~Loshchilov and F.~Hutter, ``{Decoupled Weight Decay Regularization},'' in
  \emph{Proc. Int. Conf. Learn. Representations}, 2019, pp. 1--11.

\bibitem{Mueller2016ECCV}
M.~Mueller, N.~Smith, and B.~Ghanem, ``{A Benchmark and Simulator for UAV
  Tracking},'' in \emph{Proc. Eur. Conf. Comput. Vis.}, 2016, pp. 445--461.

\end{thebibliography}

\end{document}